\documentclass[11pt]{article}
\usepackage[a4paper,margin=2.2cm]{geometry}
\usepackage{newtxtext,newtxmath,amsmath,booktabs,graphicx,siunitx,microtype}
\usepackage{xurl}
\usepackage[hidelinks]{hyperref}
\usepackage[numbers,sort&compress]{natbib}
\title{Task-Distribution-Aware Counterweight Synthesis and Constrained Co-Design for Serial Manipulators}
\author{Mohammad Abbadi\\College of Engineering and Information Technology, University of Dubai\\Dubai, United Arab Emirates\\\texttt{mabbadi@ud.ac.ae}}
\date{9 September 2026\\\small Preprint -- not peer reviewed}
\begin{document}
\maketitle
\begin{abstract}
Passive counterweights are mechanically simple gravity compensators, but a counterweight selected from a single pose is not generally optimal for the configurations and tasks a manipulator actually executes. This paper develops a task-distribution-aware synthesis framework in which the operating measure $\rho(\mathbf q)$ enters the design explicitly. For a counterweight moment $p=m_c r_c$ whose gravity torque is $-gp\phi(\mathbf q)$, the weighted mean-square residual gravity torque has the closed-form minimizer $p^*=\mathbb{E}_\rho[\tau_g\phi]/(g\mathbb{E}_\rho[\phi^2])$. If payload gravity torque is affine in payload mass, the optimum is likewise affine, $p^*(m_p,\rho)=p_0^*(\rho)+m_pK_p(\rho)$. The formulation also makes a second design fact explicit: for fixed static moment, added counterweight inertia is $I_c=pr_c$ while counterweight mass is $m_c=p/r_c$, so mass--radius selection is underdetermined unless packaging, mass, structural, or actuator constraints are supplied. A recovered three-link physical manipulator is used as a transparent case study. At $r_c=0.20$ m, equivalent zero-payload optima are 0.672 kg for uniform joint-space operation, 0.683 kg for approximately uniform task-space operation, 0.713 kg for a representative pick-and-place family, and 0.952 kg for a declared high-gravity-biased distribution. Thus the selected mass changes by more than 40\% solely through operating-distribution choice. Full nondominated fronts show that geometric knees move with declared engineering bounds, directly exposing the need for physical constraints. A rated-torque-referenced all-joint screen increases zero-payload feasible task-space coverage from 78.1\% without compensation to 93.7\% for the uniform-distribution design. A lumped point-mass trajectory study identifies a provisional crossover from no counterweight at very aggressive motion to stronger compensation as motion slows. These actuator and dynamic results are engineering consequence studies rather than physical validation; identified rigid-body inertias, friction, and paired experiments remain necessary for hardware-level claims.\end{abstract}
\noindent\textbf{Keywords:} gravity compensation; counterweight synthesis; task distribution; static balancing; serial manipulator; constrained multi-objective design.

\section{Introduction}
Passive gravity compensation reduces the actuator effort required to support links and payloads without continuous external power. Counterweights, springs, cable--pulley systems, compliant mechanisms, magnetic devices, and gear--spring modules have all been used for this purpose \cite{Arakelian2016,Nguyen2026Review,Vyas2025}. Counterweights are particularly transparent mechanically, but the mass added to generate a balancing moment also adds inertia and joint reaction. That trade-off becomes important in lightweight manipulators where the balancing mass is comparable with the moving mass \cite{Martini2019,Schneegans2024}.

Recent mechanism research has moved beyond the binary question of whether gravity can be balanced. Work has addressed four-bar and gear--spring synthesis \cite{Nguyen2022Fourbar,KuoWu2023,Hsu2025,Kuo2025Fourbar}, spatial and multi-DoF compensation \cite{Shi2023,Juang2024,Peng2025,Wang2025}, variable payloads \cite{Nguyen2025Variable}, permanent-magnet compensation \cite{Zeng2025}, and dynamic degradation under velocity, acceleration, and unbalanced loads \cite{Nguyen2024Dynamic}. Current reviews therefore identify payload variability, multi-DoF coupling, non-ideal components, and multi-objective trade-offs as active challenges \cite{Nguyen2026Review}.

For a pure counterweight, however, two design ambiguities remain easy to conceal. Static compensation is governed by the first mass moment
\begin{equation}p=m_cr_c,\end{equation}
whereas the point-mass inertia added about the compensated joint is
\begin{equation}I_c=m_cr_c^2=pr_c.\end{equation}
Infinitely many mass--radius pairs therefore generate the same static moment but have different inertia, total mass, packaging demand, and structural consequence. Equally importantly, the moment that should be compensated is not unique unless the manipulator's intended operating distribution is declared.

The central question of this paper is: \emph{for a declared operating distribution and payload model, what passive counterweight moment minimizes residual gravity torque, and what physical constraints are required to turn that moment into a unique mass--radius design?}

The paper makes four contributions. First, it derives a general operating-distribution-aware closed-form synthesis law and a payload-affine extension. Second, it formalizes why counterweight mass--radius realization is underdetermined without physical constraints. Third, it quantifies task-distribution sensitivity and replaces an arbitrary weighted-knee heuristic with complete nondominated fronts and a weight-free geometric diagnostic under explicitly labeled engineering scenarios. Fourth, it demonstrates system consequences on a recovered three-link prototype using all-joint rated-torque reference classes, quasi-static task-space feasibility, and a provisional dynamic-regime map. Historical hardware is used for provenance, not as experimental validation of the new designs.

\section{Related work and positioning}
Classical static balancing has a mature theoretical foundation for serial and parallel mechanisms \cite{Wang1999,Wang2000,Agrawal2004,Russo2005,Arakelian2016}. Modern balancing taxonomies distinguish gravity balancing from inertial and force balancing, an important distinction because a statically attractive mechanism can remain dynamically unfavorable \cite{Schneegans2024}. Martini et al. combined counterweights and springs and ranked variants using task-level indicators \cite{Martini2019}; later work introduced compact spring, cable-driven, coupled-spring, and permanent-magnet compensators \cite{Shi2023,Juang2024,Hsu2025,Wang2025,Zeng2025}.

Dynamic-load analysis has shown that the benefits of static balancing degrade with velocity, acceleration, and unbalanced payload \cite{Nguyen2024Dynamic}. Variable-payload and multi-objective formulations are likewise established \cite{Nguyen2024Multiobjective,Nguyen2025Variable,Nguyen2022Reliability}. The contribution here is therefore not the existence of counterweights, variable-payload compensation, or generic multi-objective optimization. It is the explicit introduction of the operating measure into closed-form counterweight synthesis, the resulting payload law, and the mechanism-design consequence that a static moment does not uniquely determine a physically preferred mass--radius pair.

\section{Task-distribution-aware counterweight synthesis}
\subsection{Weighted gravity-compensation objective}
Consider a serial manipulator with configuration $\mathbf q\in\mathcal Q$ and a counterweight acting about one selected joint. Let the uncompensated gravity torque about that joint be $\tau_g(\mathbf q,m_p)$, where $m_p$ is payload mass. A counterweight with moment $p=m_cr_c$ contributes
\begin{equation}\tau_c(\mathbf q,p)=-gp\phi(\mathbf q),\end{equation}
where $\phi$ is the gravity projection determined by the mounting geometry. For the shoulder-mounted case study, $\phi(\mathbf q)=\cos q_1$. Let $\rho(\mathbf q)\ge0$ be a normalized operating density. The weighted mean-square residual gravity torque is
\begin{equation}
J(p)=\int_{\mathcal Q}\rho(\mathbf q)\left[\tau_g(\mathbf q,m_p)-gp\phi(\mathbf q)\right]^2d\mathbf q.
\end{equation}

\paragraph{Proposition 1 (weighted RMS-optimal passive moment).}
If $\mathbb E_\rho[\phi^2]>0$, $J(p)$ is strictly convex and has the unique unconstrained minimizer
\begin{equation}
\boxed{p^*(m_p,\rho)=\frac{\mathbb E_\rho[\tau_g(\mathbf q,m_p)\phi(\mathbf q)]}{g\mathbb E_\rho[\phi(\mathbf q)^2]}}.
\label{eq:pstar}
\end{equation}
The result follows directly from $dJ/dp=-2g\mathbb E_\rho[(\tau_g-gp\phi)\phi]$ and $d^2J/dp^2=2g^2\mathbb E_\rho[\phi^2]>0$. Manipulator geometry and mass distribution enter through $\tau_g$; intended operation enters through $\rho$.

\subsection{Payload-affine synthesis}
For rigid-link gravity loading, payload contribution is linear in payload mass. Write
\begin{equation}\tau_g(\mathbf q,m_p)=S_0(\mathbf q)+m_pS_p(\mathbf q).\end{equation}
Substitution into Eq.~\eqref{eq:pstar} gives
\begin{equation}
\boxed{p^*(m_p,\rho)=p_0^*(\rho)+m_pK_p(\rho)},
\end{equation}
with
\begin{equation}
p_0^*=\frac{\mathbb E_\rho[S_0\phi]}{g\mathbb E_\rho[\phi^2]},\qquad
K_p=\frac{\mathbb E_\rho[S_p\phi]}{g\mathbb E_\rho[\phi^2]}.
\end{equation}
Thus the optimum is affine in payload under the stated gravity model, but both intercept and slope depend on the declared task distribution.

\subsection{Mass--radius underdetermination}
For a prescribed static moment $p$, $m_c=p/r_c$ and $I_c=pr_c$. Reducing $r_c$ reduces point inertia while increasing mass. Hence neither static torque nor inertia alone yields a unique physical pair $(m_c,r_c)$. Packaging, mass, structural, clearance, or actuator constraints are required to make the co-design well posed. This observation is used below to interpret, rather than hide, boundary-active Pareto solutions.

\section{Case-study model and verification}
\subsection{Recovered physical parameters}
The case study is a planar three-link manipulator reconstructed from original project records. The moving-link lengths are $L_1=0.23$ m, $L_2=0.20$ m, and $L_3=0.14$ m. The audited lumped masses are W2=0.214 kg, W3=0.060 kg, W4=0.060 kg, W5=0.056 kg, W6=0.126 kg, W7=0.056 kg, and W8=0.078 kg. The source counterweight is 0.887 kg at 0.20 m. Historical records identify a Power HD HD1235MG base servo and 1501MG joint servos, but their published stall ratings are not treated as continuous-duty actuator limits.

\begin{figure}[t]
\centering\includegraphics[width=\columnwidth]{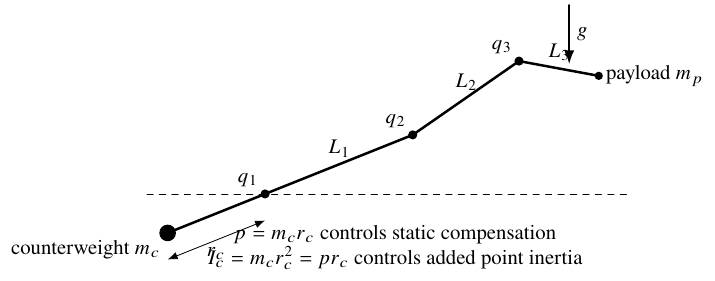}
\caption{Planar three-link case-study model. The counterweight moment $p=m_cr_c$ and its inertia $I_c=pr_c$ separate static compensation from mass--radius realization.}
\end{figure}

At the horizontal pose, the audited model gives a shoulder gravity torque of 2.17384 N m. The historical source counterweight reduces it to 0.43414 N m, an 80.03\% pose-specific reduction. This value is retained as a provenance check, not as the principal contribution.

\subsection{Gravity-vector verification}
For point mass $i$, let $\mathbf r_i(\mathbf q)$ be its planar position and $J_i$ its translational Jacobian. The reconstructed gravity vector is
\begin{equation}\mathbf G(\mathbf q)=\sum_i m_i g J_{i,y}(\mathbf q)^T.\end{equation}
The counterweight changes only the shoulder component by $-gp\cos q_1$. The implementation is checked against an independently evaluated potential-energy gradient and grid-convergence tests in the accompanying audit package. These are numerical verification tests; they are not physical validation of the robot.

\section{Operating-distribution and payload results}
Five declared operating measures are examined: uniform joint space; an approximately uniform task-space occupancy measure using a 60$\times$60 Cartesian occupancy grid; a representative family of three quintic point-to-point joint trajectories; a Gaussian high-gravity bias near extended configurations; and a folded-biased Gaussian sensitivity scenario. The latter two are declared scenarios rather than measured duty-cycle distributions.

\begin{table}[t]
\centering\small
\caption{Zero-payload task-distribution-aware synthesis at $r_c=0.20$ m.}
\begin{tabular}{lccc}
\toprule
Distribution & $p^*$ (kg m) & $m_c$ (kg) & RMS (N m)\\
\midrule
Uniform joint space & 0.13446 & 0.672 & 0.592\\
Approx. uniform task space & 0.13656 & 0.683 & 0.623\\
Pick-and-place family & 0.14259 & 0.713 & 0.554\\
High-gravity-biased & 0.19045 & 0.952 & 0.295\\
Folded-biased & 0.14401 & 0.720 & 0.698\\
\bottomrule
\end{tabular}
\end{table}

The same robot therefore admits materially different optimal moments solely because $\rho$ changes. Relative to uniform joint-space operation, the high-gravity-biased equivalent mass is more than 40\% larger. Accordingly, the phrase ``optimal counterweight'' is incomplete unless the objective measure is stated.

\begin{figure}[t]
\centering\includegraphics[width=\columnwidth]{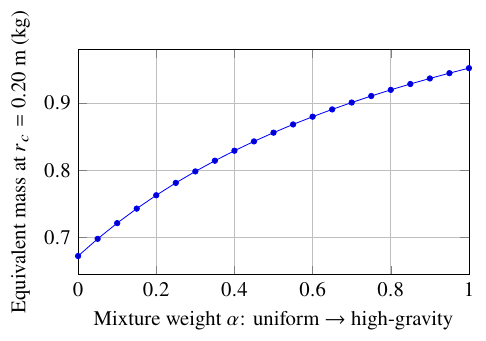}
\caption{Continuous sensitivity of the equivalent optimum to a mixture between uniform joint-space and high-gravity-biased operating measures.}
\end{figure}

The payload-affine law is numerically satisfied to machine precision over 0--0.5 kg for every tested distribution. For uniform joint-space operation at $r_c=0.20$ m,
\begin{equation}
m_c^*(m_p)\approx 0.6723 + 1.3089m_p \quad (\mathrm{kg}).
\end{equation}
The slope is distribution dependent, as predicted analytically.

\begin{figure}[t]
\centering\includegraphics[width=\columnwidth]{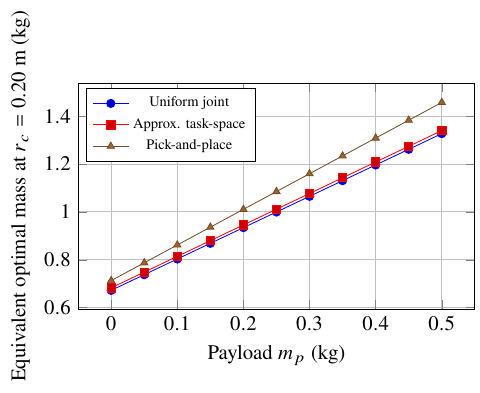}
\caption{Payload-affine equivalent optimum for three declared operating distributions.}
\end{figure}

\section{Constrained mass--radius co-design}
The previous weighted scalar knee is replaced by full nondominated fronts in residual RMS gravity torque and added counterweight inertia, with mass and radius reported explicitly. A geometric knee, when shown, is selected by maximum deviation from the chord joining normalized Pareto endpoints; no subjective mass weight enters the construction.

The historical archive establishes only one directly documented radius, 0.20 m, and does not establish structural hard limits. Accordingly, the study separates a historical fixed-radius analysis from three explicitly labeled engineering scenarios (compact, moderate, and extended). These are scenario bounds for sensitivity, not hardware-certified constraints.

\begin{figure}[t]
\centering\includegraphics[width=\columnwidth]{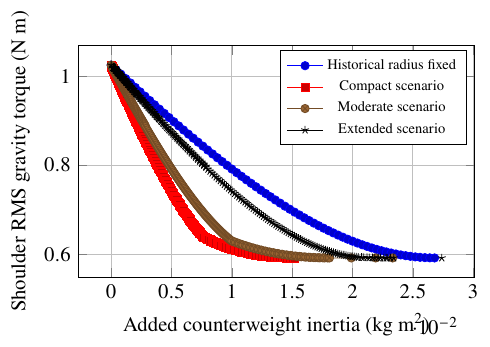}
\caption{Weight-free RMS-torque versus added-inertia nondominated fronts under declared engineering-bound scenarios. Boundary movement is itself evidence that physical packaging and structural constraints are required.}
\end{figure}

The selected geometric knee changes with the admissible design box and can become mass- or radius-bound. This is not interpreted as a failure of optimization: it numerically confirms the analytical result that a counterweight moment does not uniquely determine a physical mass--radius realization.

Parameter uncertainty is treated conservatively as sensitivity rather than reliability. Independent 2\%, 5\%, and 10\% perturbation scenarios are applied to reconstructed mass/lever contributions. Because these are not measurement-derived probability distributions, no probabilistic reliability claim is made.

\section{Actuator-reference and workspace implications}
The recovered historical hobby servos publish stall rather than defensible continuous-duty torque. For an engineering threshold study only, current CubeMars frameless motors are used as reference classes: RO60, RO80, and RO100 publish rated torques of approximately 0.8, 1.3, and 4 N m, respectively \cite{CubeMarsRO60,CubeMarsRO80,CubeMarsRO100}. They are not claimed as drop-in replacements.

Under nominal static peak torque, the uncompensated shoulder requires 2.174 N m at the horizontal pose, the source design leaves about 1.307 N m peak over the sampled envelope, and the uniform-$\rho$ design reduces the sampled peak to approximately 1.001 N m. This threshold crossing is an actuator-class feasibility indication, not sustained downsizing evidence; dynamic peaks, thermal duty, mechanical integration, and safety margins remain to be checked.

A quasi-static task-space cell is considered feasible if at least one sampled configuration reaches it while all three gravity torques satisfy reference limits of 1.3, 0.8, and 0.8 N m for joints 1--3. A zero-thickness link-1/link-3 self-intersection test removes obvious planar self-crossings. Finite link thickness and counterweight swept volume are unavailable.

\begin{figure}[t]
\centering\includegraphics[width=\columnwidth]{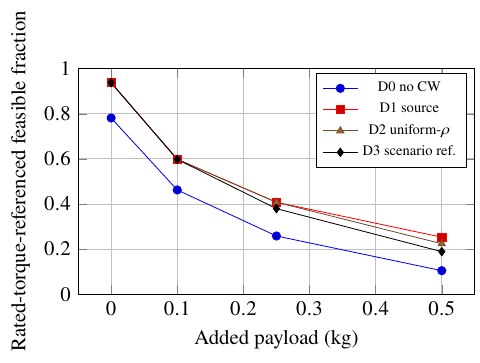}
\caption{All-joint rated-torque-referenced quasi-static task-space feasibility versus payload. The motor classes are engineering references rather than validated replacements.}
\end{figure}

At zero payload, feasible coverage increases from 78.1\% without compensation to 93.7\% for the uniform-distribution design. At 0.25 kg payload, the corresponding fractions are 25.8\% and 40.8\%. Compensation therefore enlarges the torque-feasible region, but payload exposes the unaltered requirements of joints 2 and 3.

\section{Provisional dynamic design regimes}
The static synthesis theory does not require a dynamic model, but motion aggressiveness determines whether added inertia can outweigh gravity benefit. The source archive lacks identified rigid-body rotational inertias, reflected motor/gear inertia, friction, and drivetrain efficiency. A point-mass inverse-dynamics surrogate is therefore retained only to identify qualitative regime hypotheses.

\begin{figure}[t]
\centering\includegraphics[width=\columnwidth]{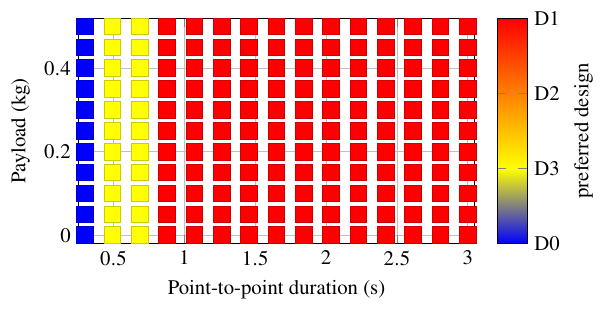}
\caption{Provisional payload--duration design-regime map from the lumped point-mass surrogate. D0: no counterweight; D1: historical source; D2: uniform-$\rho$ design; D3: declared scenario reference.}
\label{fig:regime}
\end{figure}

Four fixed designs are compared on the same quintic point-to-point family: no counterweight (D0), the historical source counterweight (D1), the uniform-joint static-RMS design (D2), and a declared scenario reference of 1.0 kg at 0.10 m (D3). Figure~\ref{fig:regime} reports the design with minimum surrogate shoulder RMS torque over payload and motion duration. Extremely aggressive motion favors no counterweight in part of the tested domain, while slower motion favors stronger compensation. The crossover is a hypothesis for identified multibody and experimental validation, not a measured motor-energy result.

\section{Generalization, limitations, and discussion}
The reusable result is Eq.~\eqref{eq:pstar}, not a prototype-specific mass. It separates a robot geometry/mass term from an operating-distribution term, while the relation $I_c=pr_c$ exposes the additional physical constraints required to realize a moment. The three-link arm demonstrates the magnitude of task dependence and the way actuator thresholds, payload, and dynamic aggressiveness can alter the preferred design.

Three evidence layers are deliberately distinguished. First, the analytical implementation is numerically verified by independent gravity/potential-energy calculations and convergence studies. Second, a recovered historical SimMechanics model establishes project provenance but is too simplified for high-fidelity validation. Third, the new counterweight designs have not been validated experimentally. The archive contains no repeated D0--D3 current, voltage, joint-torque, or measured tracking data.

The dominant remaining limitations are physical rather than numerical: identified rigid-body inertias and COMs, gearbox/motor inertia, friction/backlash and efficiency, actual mass/radius packaging bounds, bracket stress and stiffness, finite-thickness collision geometry, continuous-duty actuator integration, and paired measurements. These limitations bound the engineering consequence studies but do not alter the closed-form static synthesis result.

\section{Conclusions}
A passive counterweight cannot be called ``optimal'' independently of how a manipulator is expected to operate. Introducing the operating measure $\rho(\mathbf q)$ into the weighted residual-gravity objective yields a closed-form synthesis law and an exact payload-affine extension under the stated model. The same formulation exposes why static counterweight moment does not uniquely determine physical mass and radius: reducing radius lowers point inertia but increases mass, so packaging, structural, or actuator constraints are necessary to make the design well posed.

On the recovered three-link case study, the equivalent zero-payload design at 0.20 m shifts from 0.672 kg under uniform joint-space operation to 0.683 kg under approximately uniform task-space weighting, 0.713 kg for the representative pick-and-place family, and 0.952 kg under the high-gravity-biased scenario. Full nondominated fronts further show that representative knees shift with declared design bounds. The all-joint rated-torque-referenced screen also indicates that compensation can expand quasi-static feasible task-space, while the lumped dynamic surrogate suggests that sufficiently aggressive motion can reverse the preference. The general theory and numerical verification are reproducible; high-fidelity multibody reconstruction and paired hardware measurements remain the next step for physical validation.

\section*{CRediT authorship contribution statement}
Mohammad Abbadi: Conceptualization, Methodology, Software, Formal analysis, Investigation of archived project records, Visualization, Writing -- original draft, Writing -- review and editing.

\section*{Funding}
This research received no specific grant from funding agencies in the public, commercial, or not-for-profit sectors.

\section*{Data and code availability}
The analysis code and derived numerical data supporting the findings are available from the corresponding author and are being prepared for public archival release. The submission package contains all source files required to reproduce the manuscript figures and numerical tables; no physical measurements absent from the historical archive are represented as new experimental data.

\section*{Declaration of competing interest}
The author declares no known competing financial interests or personal relationships that could have appeared to influence the work reported in this paper.

\
\bibliographystyle{unsrtnat}
\bibliography{references}
\end{document}